\documentclass[lettersize,journal]{IEEEtran}
\usepackage{amsmath,amsfonts}
\usepackage{algorithmic}
\usepackage{algorithm}
\usepackage{array}
\usepackage[caption=false,font=normalsize,labelfont=sf,textfont=sf]{subfig}
\usepackage{textcomp}
\usepackage{stfloats}
\usepackage{url}
\usepackage{verbatim}
\usepackage{graphicx}
\usepackage{cite}
\usepackage{mathabx}
\usepackage{multirow}
\usepackage{amssymb}

\begin{document}

\title{Mutual Attention-based Hybrid Dimensional Network for Multimodal Imaging Computer-aided Diagnosis}

\author{Yin Dai, Yifan Gao, Fayu Liu, Jun Fu,~\IEEEmembership{Senior Member,~IEEE}
        % <-this % stops a space
}

\markboth{Preprint}%
{Shell \MakeLowercase{\textit{et al.}}: A Sample Article Using IEEEtran.cls for IEEE Journals}


\maketitle

\begin{abstract}
Recent works on Multimodal 3D Computer-aided diagnosis have demonstrated that obtaining a competitive automatic diagnosis model when a 3D convolution neural network (CNN) brings more parameters and medical images are scarce remains nontrivial and challenging. Considering both consistencies of regions of interest in multimodal images and diagnostic accuracy, we propose a novel mutual attention-based hybrid dimensional network for MultiModal 3D medical image classification (MMNet). The hybrid dimensional network integrates 2D CNN with 3D convolution modules to generate deeper and more informative feature maps, and reduce the training complexity of 3D fusion. Besides, the pre-trained model of ImageNet can be used in 2D CNN, which improves the performance of the model. The stereoscopic attention is focused on building rich contextual interdependencies of the region in 3D medical images. To improve the regional correlation of pathological tissues in multimodal medical images, we further design a mutual attention framework in the network to build the region-wise consistency in similar stereoscopic regions of different image modalities, providing an implicit manner to instruct the network to focus on pathological tissues. MMNet outperforms many previous solutions and achieves results competitive to the state-of-the-art on three multimodal imaging datasets, i.e., Parotid Gland Tumor (PGT) dataset, the MRNet dataset, and the PROSTATEx dataset, and its advantages are validated by extensive experiments.
\end{abstract}

\begin{IEEEkeywords}
Computer-aided diagnosis, Machine learning, Image classification, 3D convolutional neural network, hybrid dimensional network, Mutual attention module, Multimodal imaging.
\end{IEEEkeywords}

\section{Introduction}
The purpose of computer-aided diagnosis is to use computer methods to analyze medical imaging of different patients to help physicians diagnose diseases. It is an important task in computer vision with many practical applications, such as lung nodule diagnosis, diabetic retinopathy screening, COVID-19 examination, etc \cite{LN, DRS, COVID1, COVID2}. 

Recently, deep learning models designed for natural image classification \cite{ResNet} has become popular in medical image analysis \cite{S_DL1, S_DL2}. However, medical images are different from typical natural images in many aspects. For example, the region of interest is small and most of them are 3D images. Generally speaking, 3D imaging tasks in the most prominent medical imaging modes (such as CT and MRI) should be solved directly in 3D, but 3D models usually have more parameters than 2D models, so training requires more labeled data \cite{ModelsGenesis}. Compared with natural images, medical images have a small amount of data in many cases. Training a three-dimensional neural network model from scratch is difficult to converge, so the model may be under-fitting.

\begin{figure}[!t]
	\centering
	\includegraphics[width=3.3in]{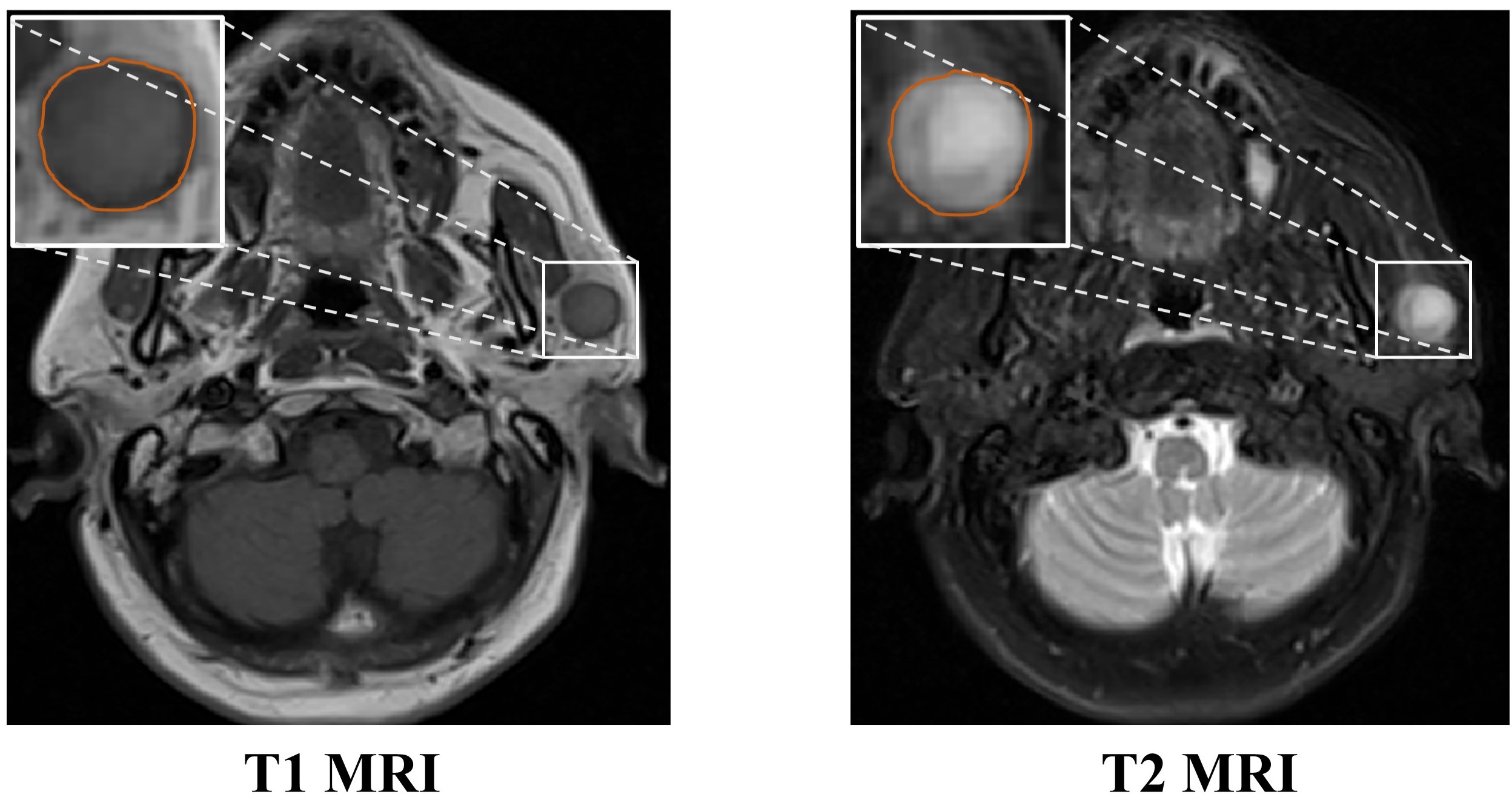}
	\caption{Examples of the slices from different MRI modalities (T1 and T2), the lesion is marked in the white box with red lines.}
	\label{fig_1}
\end{figure}

Furthermore, multimodal imaging is essential for the development of a comprehensive pathological model and has attracted increasing attention from researchers in recent years \cite{S_MMDL, Joint, HyperDenseNet, Fully, MMFNet}. In the research of medical imaging, different imaging modes are usually combined to overcome the limitations of independent imaging technology.

For example, as shown in Fig. \ref{fig_1}, in magnetic resonance imaging (MRI) \cite{MRI}, T1 images produce well contrast between human anatomical structures, while T2 images help to visualize lesions. The latest research, such as HyperDenseNet \cite{HyperDenseNet}, has been proposed to solve the problem of multimodal 3D medical image recognition. Nevertheless, these models have a large number of parameters. Due to the small region of interest in medical images, there are many redundant and unnecessary calculations.

Therefore, we design a transferable, lightweight mutual attention-based hybrid dimension network to effectively improve the accuracy of multimodal 3D medical image classification.

\begin{figure*}[!t]
	\centering
	\includegraphics[width=7in]{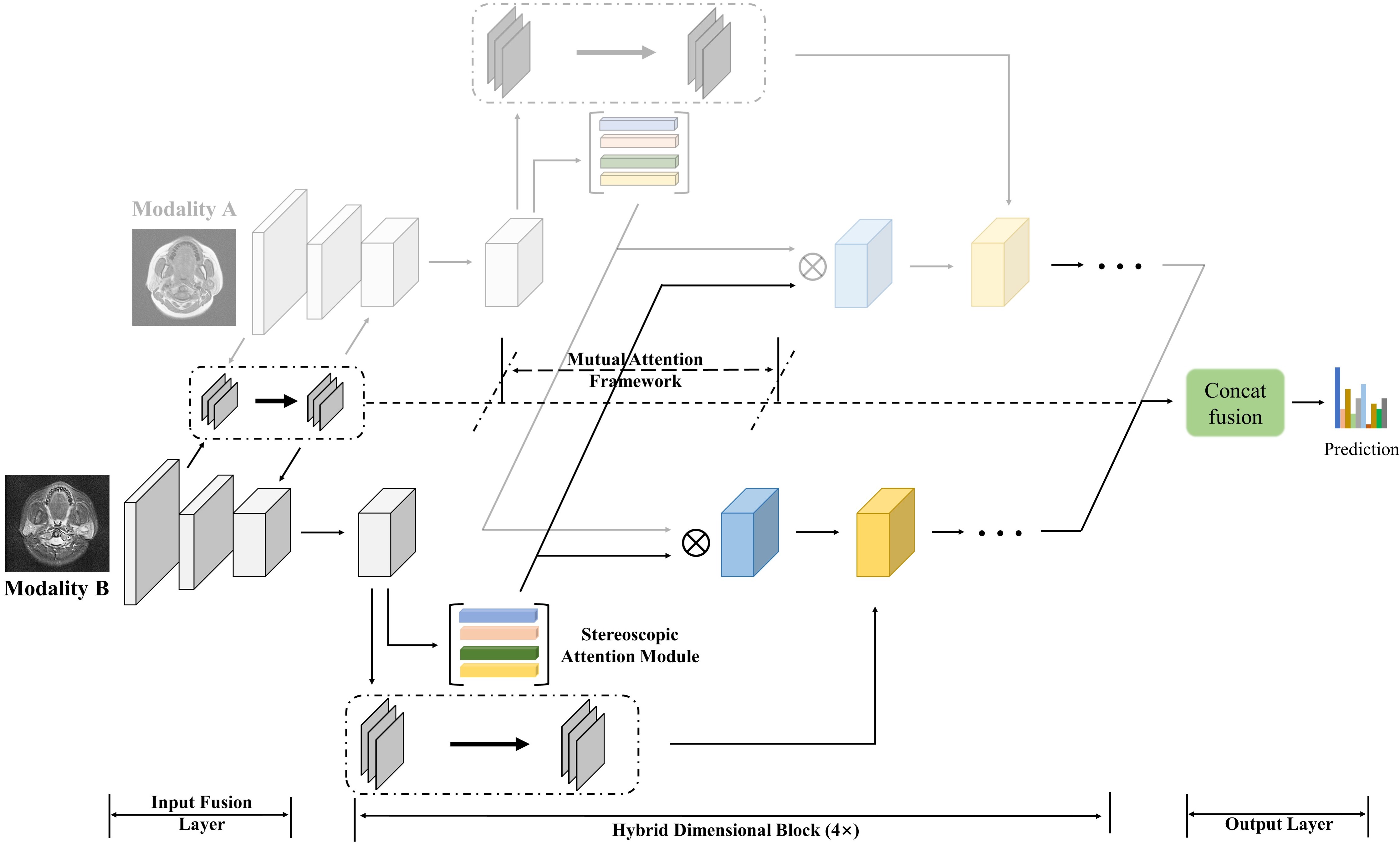}
	\caption{An overview of the proposed Mutual Attention-based Hybrid Dimensional Network. It consists of hybrid dimensional blocks, the stereoscopic attention module, and the mutual attention connection. Multiple hybrid dimensional blocks are stacked in the neural network to improve the feature extraction ability of the model.}
	\label{fig_2}
\end{figure*}

Specifically, our proposed network consists of three key components, Hybrid Dimensional Block (HDB), Stereoscopic Attention Module (SAM), and Mutual Attention Framework (MAF). HDB aims to combine 3D convolutions and 2D convolutions to reduce network parameters, and use pre-trained models from ImageNet \cite{ImageNet} in 3D images to improve performance. SAM is a novel attention module, which captures the pathological information of changing continuously at different depths in the same region. MAF transfers SAM between the dual deep network. Because the important abnormal information of images with different modalities is located in similar areas of images, it can strengthen the information interaction and augment the consistency of pathological detection between the dual deep network. In summary, we make the following four contributions:

\begin{enumerate}
	\item{We present a new architecture to improve the performance of multimodal imaging computer-aided diagnosis with our proposed hybrid dimensional network. It allows the use of pre-trained models from ImageNet in 3D images and carries much fewer parameters than the existing 3D models.}
	\item{We propose a novel stereoscopic attention module in this work, which can be leveraged to capture contextual information from region-wise dependencies in a more efficient way.}
	\item{A mutual attention connection is incorporated for passing efficient high-level feature representations of spatial information within the network. It aims at preserving multi-modality image consistency through leveraging feature recalibration.}
	\item{4. Experimental evaluations demonstrate that the proposed method achieves the most advanced performance on the in-house Parotid Gland Tumor (PGT) dataset, the MRNet dataset, and the PROSTATEx dataset.}
\end{enumerate}

The rest of this paper is organized as follows. Section II presents some closely related works. The pipeline of our proposed method is in Section III. Section IV introduces the experimental results and details. Finally, we summarize our work in Section V.

\section{Related Work}
\subsection{Hybrid Dimensional Network}
The mixed-dimensional network aims to reduce the calculations in 3D convolutional neural networks and uses pre-trained models from 2D image datasets (such as ImageNet). In video action recognition, R2D \cite{R2+1D} is a simple method that combines depth and channel and sends it to a 2D residual network for calculation. Pseudo-3D residual networks (P3D) \cite{P3D} and R(2+1)D \cite{R2+1D} use 2D spatial convolution and 1D temporal convolution to approximate 3D convolution and initialize with model parameters pre-trained on ImageNet. However, in computer-aided diagnosis, since diseased tissues are often gathered in a three-dimensional area, although 1D temporal convolution can reduce parameters, it will lose a large amount of stereoscopic context information of the pathological area. Therefore, we propose a structure that combines 2D convolution and 3D convolution, which not only preserves the unique stereoscopic information of medical images, but also reduces the parameters reasonably.

\subsection{Attention Mechanism}
The attention mechanism \cite{S_Attention} is an effective technology, which can help the model pay more attention to important information. In recent years, the attention mechanism has made important breakthroughs in image processing, natural language processing \cite{NLP}, and other fields, and has been proved to be beneficial to improve the performance of the model. SE-Net \cite{SENet} proposed for the first time an effective mechanism to learn channel-wise attention and obtain competitive performance. Non-local neural networks \cite{Non-local} generate an attention map through the correlation matrix between each point in the feature map, and then the attention guides the aggregation of rich contextual information. CBAM \cite{CBAM} is a simple yet effective attention module that uses spatial and channel attention to improve performance. ECA-Net \cite{ECANet} made some improvements to SE-Net, and proposed a local cross-channel interaction strategy without dimensionality reduction. DRA-Net \cite{DRANet} adaptively capture contextual information based on the relation-aware attention mechanism.

In the clinical diagnosis task, the radiologists first selectively pays attention to the abnormal area, and then examines it in detail. Inspired by this human visual attribute, many papers use attention-based deep learning methods to highlight the possible lesions in the image. \cite{Skin_Attention} proposed an attention residual learning convolutional neural network model for skin lesion classification in dermoscopy images. \cite{Breast_Attention} is a deep selective attention network for breast cancer classification. \cite{Self_Attention} captures richer contextual dependencies based on the use of guided self-attention mechanisms.

Although the attention mechanism can effectively improve the performance of deep networks in large training datasets, using attention weights with a large number of additional parameters may not only lead to computational costs, but also overfitting small scale training datasets. In comparison, our method only adds a very small number of parameters, and takes into account the interaction of three-dimensional region and the relevant attention of different modal images.

\subsection{Multimodal Medical Image Classification}
Multimodal medical classification is the most fundamental and challenging part in medical image analysis. It is proved that a reasonable fusion of different modalities has been a potential means to enhance Deep CNNs. Multimodal fusion can capture more abundant abnormal information and improve the quality of diagnosis.

\cite{Joint} sets three modality-specific encoders to capture low-level features and a decoder to fuse low-level and high-level features. HyperDenseNet \cite{HyperDenseNet} builds dual deep networks for different modalities of MRI and links features across these streams. \cite{Fully} fuses final features from modality-specific paths to make final decisions. MMFNet \cite{MMFNet} provides a more complex structure to fuse multimodal MRI images.

In particular, HyperDenseNet demonstrates the potential of dual deep network and cross-modal information fusion in improving model performance. On the other hand, in 3D medical image recognition, cross-modal feature maps are computationally expensive and redundant. Therefore, based on a dual deep network, we propose a method of attention weight fusion. By transferring the attention weight between different modes instead of the feature map, the network can efficiently capture complementary information.

\subsection{Transfer learning in 3D Medical Image}
Transfer learning \cite{S_TL} uses a powerful pre-trained network as a feature extractor, which is an efficient paradigm to improve the performance of the model. In fact, in medical image computing, fine-tuning ImageNet's pre-trained model has become a common method. However, the pre-trained model of ImageNet requires that the input of the neural network model must be two-dimensional images and cannot be used on three-dimensional images. For three-dimensional medical images, there is currently no pre-trained model that can generally perform transfer learning. At present, the existing migration models suitable for 3D medical imaging include MedicalNet \cite{MedicalNet} and Model Genesis \cite{ModelsGenesis}. MedicalNet trains the model in several open source medical datasets. Model Genesis obtains the pre-trained model of 3D medical images through self-supervised learning without using labels. Furthermore, MMFNet \cite{MMFNet} proposes an initialization strategy named self-transfer, which trains several modality-specific encoder-decoder models respectively, and these pre-trained encoders will be used as the initial encoders for multi-modality model. On the contrary, due to the different imaging methods of medical images and the constraints of the network structure, they cannot be well migrated to downstream tasks.

Our method combines the 2D residual network into the 3D convolutional neural network so that the 3D convolutional neural network can load pre-trained weights from ImageNet. Besides, it can be simply embedded into other networks to improve the performance of the model.

\section{Proposed Network}
\subsection{Overview of Our Method}
It is well known that multimodal MRI images carry a lot of anatomical and pathological information. This urges us to build a classification model based on the dual deep network, making full use of the complementary information of two kinds of information sources.

In this article, we propose a novel network that aims to improve feature representation for multimodal medical image classification. As shown in Fig. \ref{fig_2}, it consists of three main components: hybrid dimensional blocks, the stereoscopic attention module, and the mutual attention connection. We use two ResNet \cite{ResNet} models as two parts of the network and load the parameters pre-trained on ImageNet. This makes the model have the ability to extract features at the very beginning of training.

\begin{figure}[!t]
	\centering
	\includegraphics[width=3.3in]{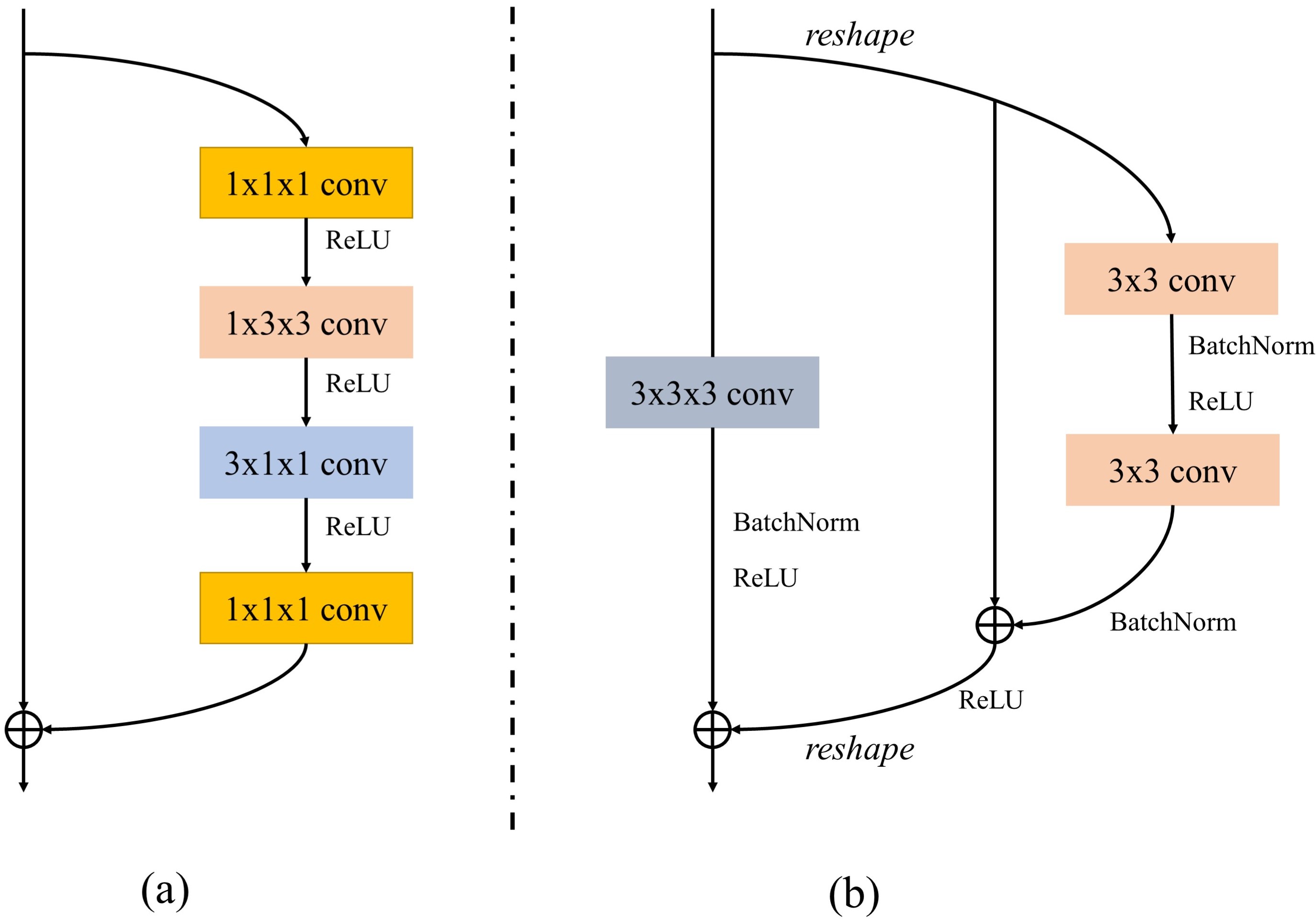}
	\caption{Details of (a) P3D-A and (b) HDB. $\oplus$ represents element-wise sum operation.}
	\label{fig_3}
\end{figure}

\subsection{Hybrid Dimensional Block}
Inspired by the successes of Pseudo-3D (P3D) in numerous challenging video classification tasks, we develop a new architecture of building modules named HDB to merge 2D convolution and 3D convolution, pursuing volume-wise encoding in an efficient way for 3D medical image classification.

The overall structure of the hybrid dimensional block is shown in Fig. \ref{fig_3}(b). Compared with the ordinary residual network, the input and output of the hybrid dimensional block are 3D images, and the main operation part is composed of 2D residual blocks instead of 3D residual blocks. Both convolutions are parallel to each other on different paths. The 3D convolution part consists of 3×3×3 3D convolution, batch normalization \cite{BN}, and ReLU \cite{ReLU}, while the 2D convolution part is a classical residual block. For intuitive understanding, we assume that the channel size is $C$, the depth is $D$, and the feature map size is $(H,W)$, then the input shape is $(C,D,H,W)$. The first 3×3×3 3D convolution has stride 1 and has $C_1$ kernels, so the shape of the output is $(C_1,D,H,W)$. The first 2D convolution in residual block requires 3D input instead of 4D input. By stacking $C_1$ channels along with the depth, the input is reshaped to $(C_1 \times D,H,W)$. After a 2D residual block, the output has a shape $(C_1 \times D,H_1,W_1)$. Finally, the channel and depth are split to obtain shape $(C_1,D,H_1,W_1)$.

\subsection{Stereoscopic Attention Module}

\begin{figure*}[!t]
	\centering
	\includegraphics[width=7in]{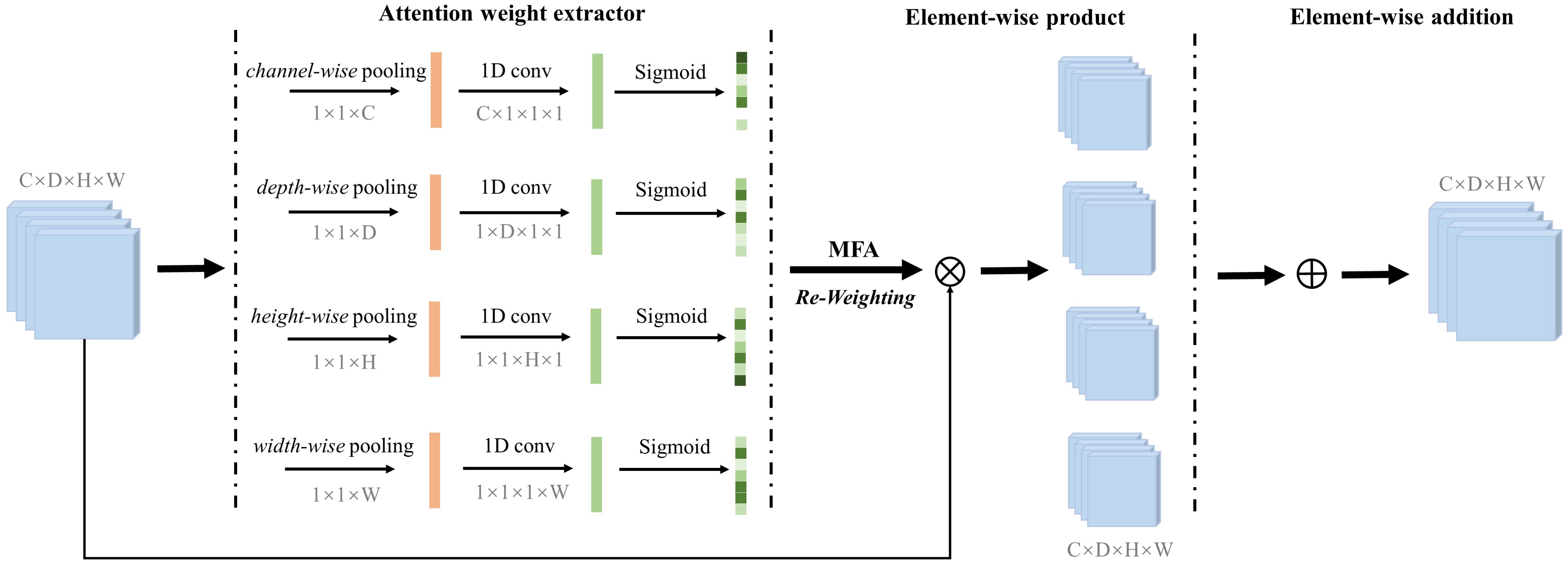}
	\caption{Overall architecture of the Stereoscopic Attention Module. $ \otimes $ represents elementwise product operation.}
	\label{fig_4}
\end{figure*}

Usually, the pathological part of a medical image changes continuously at different depths in the same region, and capturing such regions is very helpful for the prediction ability of the model. However, the current research on spatial attention mainly focuses on the relationship between pixels rather than region \cite{Non-local, ECANet, DRANet}, and medical image recognition is not highly dependent on related information between pixels. In addition, the spatial attention mechanism brings a lot of computational complexity and parameters, which not only makes training more difficult but also improves the difficulty of model optimization.

To model the context of the region in 3D medical images, we propose the stereoscopic attention module. Specifically, we first define $i$ as the set of four dimensions:

\begin{equation}
	i=\{C,D,H,W\}
\end{equation}

Then, we directly construct the weighted relationship between regions. As shown in Fig. \ref{fig_4}, given a feature $X\in R^{C \times D \times H \times W}$, where $C$, $D$, $H$ and $W$ are channel, depth, height and width. First, global average pooling (GAP) is applied to the four dimensions of channel, depth, height and width, and four features $X_{GAP}^C$, $X_{GAP}^D$, $X_{GAP}^H$ and $X_{GAP}^W$ are generated.

\begin{equation}
	X_{GAP}^i = f_{GAP}^i(X)
\end{equation}

Next, we reshape the four features into $(1,1,C)$, $(1,1,D)$, $(1,1,H)$, $(1,1,W)$, and send them to four different one-dimensional convolutional layers respectively. After that, they are activated by the sigmoid function, and the results are reshaped into $(C,1,1,1)$, $(1,D,1,1)$, $(1,1,H,1)$, $(1,1,1,W)$.

\begin{equation}
	X_{Conv}^i=Conv1D(f_{reshape}(X_{GAP}^i))
\end{equation}

\begin{equation}
	X_{aw}^i=f_{reshape}(Sigmoid(X_{Conv}^i))
\end{equation}

We get the four-dimensional attention weight vector $X_{aw}^C$, $X_{aw}^D$, $X_{aw}^H$, $X_{aw}^W$, and perform elementwise product between $X$. 

\begin{equation}
	X^i=X_{aw}^i \otimes X
\end{equation}

Finally, perform elementwise sum operation with them to get the final output $O\in R^{C \times D \times H \times W}$ as follows: 

\begin{equation}
	O=X^C \oplus X^D \oplus X^H \oplus X^W
\end{equation}

SAM explicit guidance model to build attention to three-dimensional area. Each attention weight interactives the information of the adjacent area through one-dimensional convolution and obtains stereoscopic perception when reconstructing from 1D to 4D.

\begin{figure}[!t]
	\centering
	\includegraphics[width=3.3in]{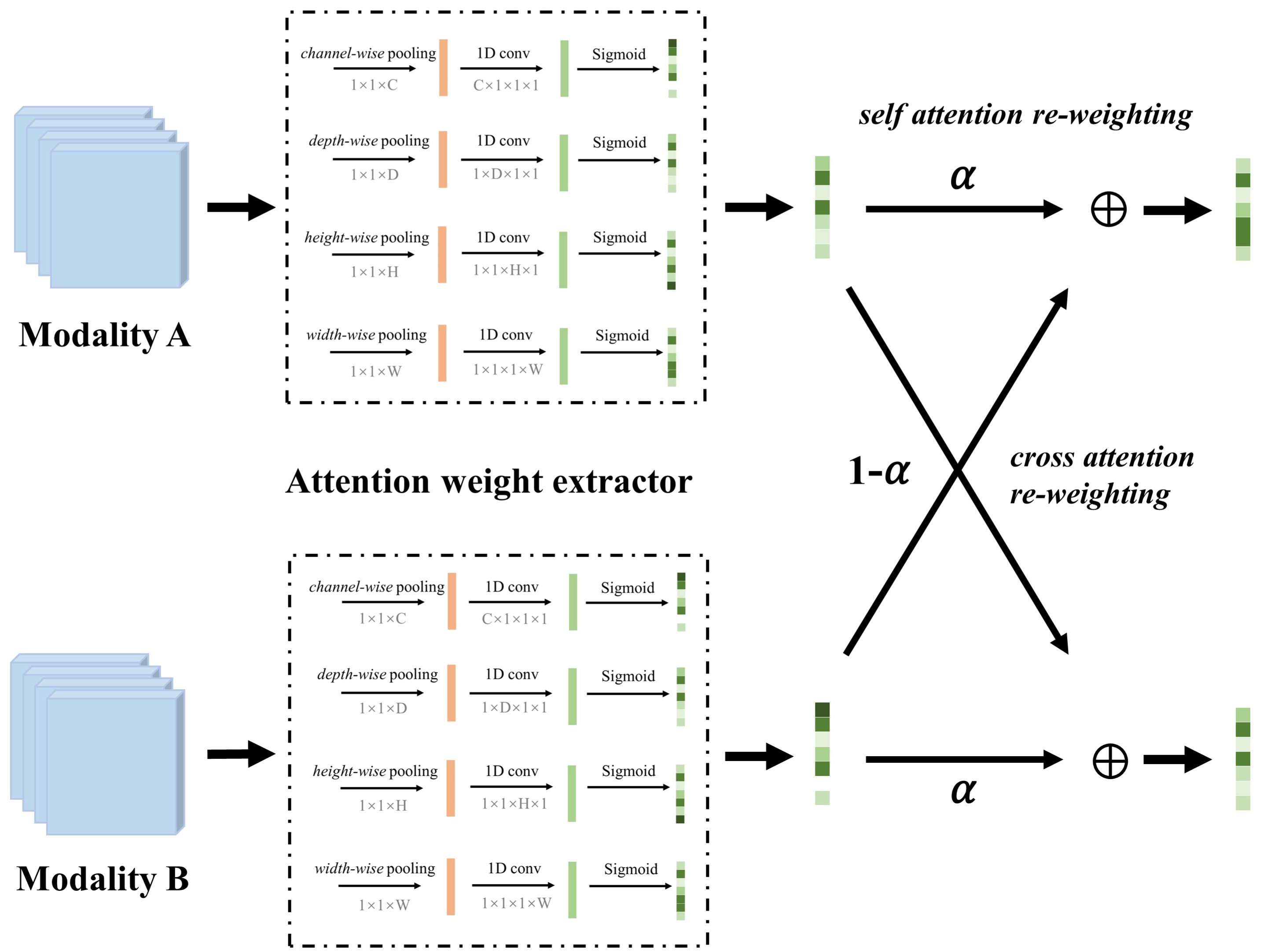}
	\caption{Overall architecture of Mutual Attention Framework. The details of attention weight extractor are shown in Fig. \ref{fig_4}. Note that for the convenience of demonstration, only one attention weight (actually four) of the output of the attention weight extractor is drawn.}
	\label{fig_5}
\end{figure}

\subsection{Mutual Attention Framework}
In medical imaging diagnosis, cross-modality images bring richer information. However, simply inputting different modalities into different backbones for calculations will lose a lot of complementary information, and this information has a significant effect on accurately diagnosing diseases.

In HyperDenseNet, the author uses the connection structure of feature maps to improve the flow of gradients and the fusion of information, but it brings a lot of parameters and unnecessary calculations. To improve this situation, we propose the mutual attention framework, which only allows attention parameters to be transferred across modalities. The mutual attention framework divides attention into self-attention and cross-attention. In the general attention mechanism, the feature map is weighted by the attention parameters from its own modality, while in the mutual attention framework, the feature map is weighted by the attention parameters of its own modality and cross-modality at the same time. This not only strengthens the flow of information between different modalities but also implicitly synthesizes complementary information between different modalities. At the same time, mutual attention also imitates the doctor's observation of different modal images, that is, looking for as much information as possible in different images to get the most reliable diagnosis.

As shown in Fig. \ref{fig_5}, we first define j and k as the set of four dimensions in self modality and cross modality, respectively.

\begin{equation}
	j=\{C_{self},D_{self},H_{self},W_{self}\}
\end{equation}

\begin{equation}
	k=\{C_{cross},D_{cross},H_{cross},W_{cross}\}
\end{equation}

Second, define the attention weight of the self-modality as $W_{self}^i$, and the attention weight of the cross-modality as $W_{cross}^i$. From Eq. 4, self-attention weight and cross-attention weight can be obtained as follows:

\begin{equation}
	W_{self}^i=X_{aw}^j
\end{equation}

\begin{equation}
	W_{cross}^i=X_{aw}^k
\end{equation}

Where $W_{self}^i$ is self-attention weight, $W_{cross}^i$ is cross-attention weight. Therefore, considering mutual attention, $X^i$ in Eq 5 is updated as follows:

\begin{equation}
	X^i=\alpha W_{self}^i \otimes X + (1-\alpha)W_{cross}^i \otimes X
\end{equation}

The factor $\alpha \in (0,1)$ determines the proportion of self-attention weight. When $\alpha$=1, the feature map is only weighted by the self-attention weight, when $\alpha$=0, the feature map is only weighted by the cross-attention weight. A reasonable value of $\alpha$ should be between 0 to 1. We will specifically discuss the effect of $\alpha$ on the model in Section IV.

\section{Results and Discussion}
To evaluate the proposed method, we conducted comprehensive experiments on the parotid gland tumor (PGT) dataset, the MRNet dataset, and the PROSTATEx dataset. Experimental results show that MMNet has reached the most advanced performance on all datasets. Next, we first introduced the datasets and implementation details and then performed a large number of ablation experiments on the parotid gland tumor dataset. Finally, we compare our approach with some of the state-of-the-art technologies on three datasets.

\begin{table*}[]
	\centering
	\caption{Mean accuracy and per-class precision on the PGT dataset. MMNet outperforms existing approaches and achieves 90.0\% in mean accuracy. * means that the model is pretrained in a large number of medical image datasets. PA means pleomorphic adenoma, MT means Warthin tumor, BCA means basal cell adenoma, and OBL means other benign lesions.}
	\label{tab1}
	\renewcommand\arraystretch{1.3}
	\begin{tabular}{llllllllll}
		\hline
		Methods                & Pre-train & Params & Flops ($10^{12}$) & Mean Acc     & PA                & WT                & MT                 & BCA                & OBL               \\ \hline
		R2D {[}12{]}  & Random   & 22M    & 0.05                          & 65.2±5.8          & 49.0±32.2         & 72.1±7.7          & 70.4±7.9           & 62.3±23.0          & 64.1±14.5         \\
		P3D-199 {[}13{]}       & Random   & 65M    & 0.77                          & 72.2±5.5          & 63.5±18.3         & 75.6±10.4         & 73.4±18.7          & 83.2±16.3          & 69.6±11.5         \\
		P3D-199 {[}13{]}       & ImageNet & 65M    & 0.77                          & 82.1±3.1          & 75.5±18.6         & 88.2±6.0          & 83.2±6.8           & 67.3±14.0          & 85.5±10.6         \\
		C3D {[}41{]}           & Random   & 28M    & 3.4                           & 71.0±4.1          & 68.3±38.9         & 81.3±15.4         & 67.8±10.0          & 71.4±7.5           & 84.5±12.4         \\
		3D Resnet18 {[}5{]}    & Random   & 33M    & 1.46                          & 71.0±3.3          & 58.3±33.3         & 70.7±9.3          & 74.0±8.5           & 78.9±20.0          & 73.2±6.7          \\
		3D Resnet34 {[}5{]}    & Random   & 64M    & 2.7                           & 73.3±5.1          & 69.2±16.2         & 86.8±5.2          & 75.3±18.1          & 68.7±13.8          & 65.7±6.4          \\
		MedicalNet {[}29{]}    & *        & 46M    & 2                             & 73.5±4.2          & 61.6±15.7         & 81.1±7.7          & 76.0±12.4          & 59.2±11.9          & 78.1±8.1          \\
		Model Genesis {[}30{]} & *        & 20M    & 18.18                         & 75.6±3.6          & 73.0±14.0         & 80.1±11.5         & 77.9±7.9           & 78.6±13.1          & 67.9±3.4          \\
		HyperDenseNet {[}27{]} & Random   & 10M    & \textgreater{}20              & 74.2±2.9          & 76.0±24.8         & 86.2±10.0         & 72.9±16.3          & 75.2±24.6          & 80.0±15.2         \\ \hline
		MMNet18 (Ours)         & Random   & 32M    & 0.49                          & 78.3±2.2          & 68.7±14.3         & 86.4±8.6          & 87.2±9.5           & 78.0±17.9          & 74.3±16.3         \\
		MMNet18 (Ours)         & ImageNet & 32M    & 0.49                          & 85.4±2.8          & \textbf{93.0±7.3} & 84.5±7.1          & 89.2±5.0           & 82.1±15.6          & 85.2±6.9          \\
		MMNet34 (Ours)         & Random   & 52M    & 0.73                          & 80.5±1.7          & 81.6±16.8         & 84.0±5.3          & 72.5±9.8           & 82.6±16.1          & 84.6±5.2          \\
		MMNet34 (Ours)         & ImageNet & 52M    & 0.73                          & \textbf{90.0±3.2} & 91.3±11.8         & \textbf{91.8±4.6} & \textbf{89.5±10.5} & \textbf{94.1±10.5} & \textbf{88.6±8.8} \\ \hline
	\end{tabular}
\end{table*}

\subsection{Dataset}
\subsubsection{PGT Dataset}
The incidence of malignant tumors in parotid gland tumors \cite{PGT1} is about 20\%. Correct preoperative diagnosis of these tumors is essential for proper surgical planning. Among them, imaging examination plays an important role in determining the nature of parotid gland masses. Magnetic resonance imaging (MRI) is considered to be the preferred imaging method for preoperative diagnosis of parotid tumors \cite{PGT2}. MRI can provide information about the exact location of the lesion, the relationship with the surrounding structure, and can assess the spread of nerves and bone invasion. However, it is reported that parotid gland tumors show considerable overlap in imaging features (such as tumor margins, homogeneity, and signal intensity), so it is difficult for doctors to identify the mass.

According to common clinical classifications, we divide parotid gland tumors into five categories: pleomorphic adenoma, Warthin tumor, malignant tumor, basal cell adenoma, and a few other benign lesions. A total of 375 patients with parotid gland lesions were studied, and some patients lacking MRI T1 or T2 images were removed, and finally, 344 patients with parotid gland lesions were included.

First, perform OTSU \cite{OTSU} and manual adjustment to extract the foreground area in the original image. Then the images of different modalities of the same patient are registered to improve the consistency of the foreground area. Then resample each image to (18, 224, 224). Therefore, 344 images are finally included, each of which is a stack of 3D images of MRI T1 and T2, and the size is (36, 224, 224). Data augmentation uses random flipping and random noise. Random flipping performs flipping of the image with 50\% probability. Random noise adds Gaussian noise with a mean value of 0 and a variance of 0.25 to the image. The patients were randomly divided into training group (n = 275) and independent test group (n = 69) according to the ratio of 4:1, and then the training group was used to optimize the model parameters.

\subsubsection{MRNet Dataset}
The MRNet dataset \cite{MRNet} is a publicly available medical image benchmark containing 1370 knee MRI examinations performed at Stanford University Medical Center. Each sample is marked for the anterior cruciate ligament (ACL), the meniscus, or other abnormal signs (Abnormality) of the corresponding knee joint. The dataset is randomly divided into 1130 training samples, 120 validation samples, and 120 test samples. The provided dataset has three MRI modalities, including T1 weighted image, T2 weighted image, and proton density weighting. In order to make a fair comparison with the baseline model, in this paper, we preprocess the data by implementing the same strategy used in MRNet. The data augmentation method is consistent with the approach used in the PGT dataset.

\subsubsection{PROSTATEx Dataset}
The PROSTATEx dataset \cite{PROSTATEx} is a training set from the SPIE-AAPM-NCI PROSTATEx challenge. It includes the imaging dataset of 204 patients and a total of 330 biopsy-proven prostate lesions (76 clinically significant and 254 clinically insignificant). Furthermore, this study excluded the remaining test data from 140 patients because their labels were publicly unavailable. The MRI protocol of the PROSTATEx dataset is described in \cite{PROSTATEx}.

\subsection{Experimental Setups and Evaluation Criteria}
We set SGD \cite{SGD} as the optimizer with a learning rate equal to  $10^{-3}$ and momentum equal to 0.9. The maximum training round is set to 100. Our experiments were performed on NVIDIA 3080 GPU (with 10GB GPU memory). The code is implemented using Pytorch \cite{PyTorch} and TorchIO \cite{TorchIO}.

To ensure a fair comparison of the experiments, we use different evaluation criteria in the public and private datasets. Among them, the evaluation criteria in the public dataset are consistent with the existing solutions. In the MRNet dataset and PROSTATEx dataset, we evaluated using accuracy, sensitivity, specificity, and area under the receiver operating characteristic curve (ROC-AUC). In the PGT dataset, The evaluation criteria for each model are the overall accuracy and the precision of each category.

\subsection{Baseline Methods}
\subsubsection{PGT Dataset}
In order to compare with the hybrid dimensional block, we select P3D and R2D as the baseline model. 3D-Resnet and C3D \cite{C3D} are included in the comparison as common 3D neural network models. HyperdenseNet and our method have a similar idea, that is, using the fusion of multimodal to calculate medical images. The reason for selecting MedicalNet and Model Genesis as baseline models is that they are both large-scale pre-trained models designed for medical images.

\subsubsection{MRNet Dataset}
In this experiment, we compare our model with three state-of-the-art implementations. MRNet \cite{MRNet} mainly consists of three AlexNets \cite{AlexNet} backbone networks. They predict each modality independently and fuse the decisions of each backbone network to derive the final diagnosis. ELNet  \cite{ELNet} mainly uses the Resnet model and proposes two techniques, multi-slice normalization and BlurPool layer, to further improve the performance. MRPyrNet \cite{MRPyrNet} uses a feature pyramid network to improve the ability to capture injuries that occur in the knee region. This module was inserted into MRNet and ELNet and significant performance improvements were achieved.

\begin{figure*}[!t]
	\centering
	\includegraphics[width=7in]{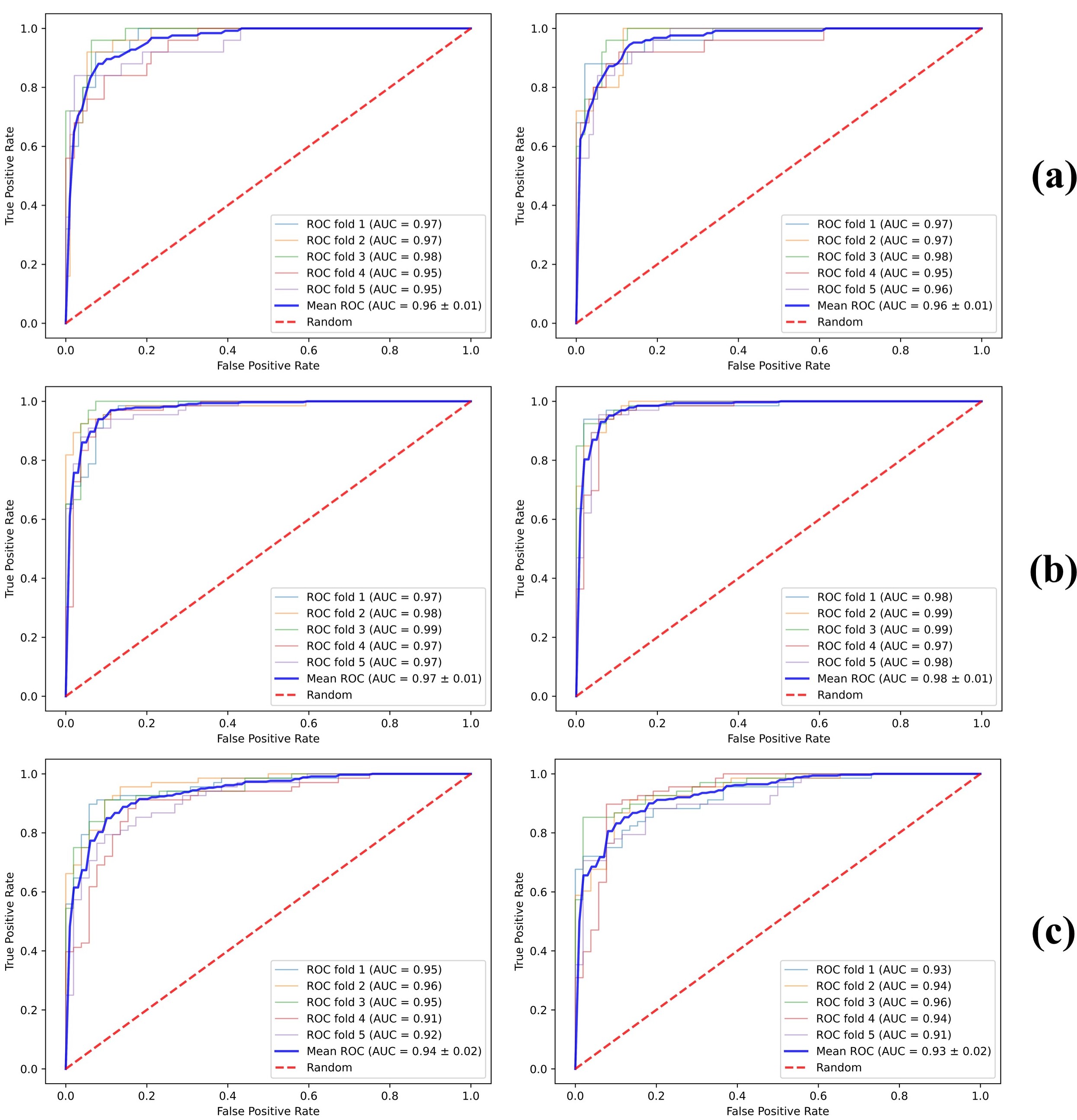}
	\caption{ROC curves for each experiment in the MRNet dataset. (a) to (c) are the results of MMNet experiments in Abnormality, ACL Tear, and Meniscus Tear, respectively. On the left is the ROC curves for MMNet18, and on the right is the ROC curves for MMNet34. Where the thin line represents the results of per-fold cross-validation and the thick blue line represents the average results.}
	\label{fig_7}
\end{figure*}

\subsubsection{PROSTATEx Dataset}
Liu et al. \cite{XmasNet} proposed XmasNet, a convolutional neural network-based, end-to-end PCa classification framework. Seah et al. \cite{Detection} used auto-window techniques and transfer learning to further improve the performance of deep models for PCa classification. Chen et al. \cite{Transfer} built deep models using InceptionV3 \cite{InceptionV3} and VGG-16 \cite{VGG} for PCa diagnosis and used pre-trained models from ImageNet to alleviate the problem of insufficient medical image data. \cite{Semi} developed a novel semi-automatic classification framework based on 3D convolutional neural networks using different MRI sequences as inputs. \cite{MISN} proposed a multi-input selection network (MISN) for PCa classification. Compared with other methods, MISN is a multiple input and output deep network for maximizing the use of multiparametric MR images from the PROSTATEx dataset to improve diagnostic performance.

\subsection{Experimental results}
\begin{table*}[]
	\centering
	\caption{Comparison on the MRNet dataset.}
	\label{tab2}
	\renewcommand\arraystretch{1.3}
	\begin{tabular}{cccccc}
		\hline
		Pathology                      & Method               & ROC-AUC              & Accuracy             & Sensitivity          & Specificity          \\ \hline
		\multirow{6}{*}{Abnormality}   & MRNet                & 0.936                & 0.883                & 0.947                & 0.64                 \\
		& ELNet                & 0.941                & 0.917                & \textbf{0.968}       & 0.72                 \\
		& MRPyrNet(with MRNet) & ——                   & ——                   & ——                   & ——                   \\
		& MRPyrNet(with ELNet) & ——                   & ——                   & ——                   & ——                   \\
		& MMNet18 (Ours)       & 0.962±0.013          & 0.915±0.006          & 0.958±0.009          & 0.752±0.053          \\
		& MMNet34 (Ours)       & \textbf{0.967±0.011} & \textbf{0.923±0.014} & 0.958±0.020          & \textbf{0.792±0.016} \\ \hline
		\multirow{6}{*}{ACL Tear}      & MRNet                & 0.955±0.005          & 0.847±0.005          & 0.722±0.000          & 0.950±0.009          \\
		& ELNet                & 0.940±0.001          & 0.808±0.000          & 0.648±0.019          & 0.939±0.015          \\
		& MRPyrNet(with MRNet) & 0.976±0.003          & 0.886±0.010          & 0.815±0.019          & 0.944±0.009          \\
		& MRPyrNet(with ELNet) & 0.960±0.015          & 0.881±0.034          & 0.827±0.039          & 0.924±0.030          \\
		& MMNet18 (Ours)       & 0.976±0.006          & 0.923±0.019          & 0.885±0.039          & 0.955±0.014          \\
		& MMNet34 (Ours)       & \textbf{0.980±0.007} & \textbf{0.937±0.009} & \textbf{0.904±0.022} & \textbf{0.964±0.007} \\ \hline
		\multirow{6}{*}{Meniscus Tear} & MRNet                & 0.843±0.016          & 0.778±0.027          & 0.750±0.067          & 0.799±0.009          \\
		& ELNet                & 0.869±0.031          & 0.775±0.044          & 0.814±0.109          & 0.745±0.075          \\
		& MRPyrNet(with MRNet) & 0.889±0.006          & 0.808±0.008          & 0.853±0.048          & 0.775±0.052          \\
		& MRPyrNet(with ELNet) & 0.895±0.008          & 0.761±0.042          & 0.872±0.106          & 0.676±0.149          \\
		& MMNet18 (Ours)       & \textbf{0.938±0.022} & \textbf{0.863±0.025} & \textbf{0.896±0.029} & 0.838±0.025          \\
		& MMNet34 (Ours)       & 0.936±0.015          & 0.860±0.021          & 0.885±0.034          & \textbf{0.841±0.017} \\ \hline
	\end{tabular}
\end{table*}

\begin{table}[]
	\centering
	\caption{Comparison on the PROSTATEx dataset.}
	\label{tab3}
	\renewcommand\arraystretch{1.3}
	\begin{tabular}{lllll}
		\hline
		& ROC-AUC    & Sensitivity & Specificity & Accuracy   \\ \hline
		Semi {[}50{]} & 0.81±0.04 & 0.70±0.06  & 0.73±0.02  & 0.63±0.02 \\
		Transfer {[}51{]}  & 0.79±0.04     & 0.75±0.07      & 0.73±0.03      & 0.63±0.03     \\
		Detect {[}52{]}  & 0.82±0.05     & 0.74±0.07      & 0.75±0.05      & 0.66±0.06     \\
		Prostate {[}53{]}     & 0.82±0.04     & 0.79±0.08      & 0.76±0.06      & 0.68±0.07     \\
		Select {[}54{]}  & 0.89±0.03     & 0.79±0.09      & 0.79±0.05      & 0.71±0.05     \\
		3D ResNet                              & 0.84±0.05     & 0.82±0.08      & 0.72±0.05      & 0.63±0.06     \\
		P3D                                    & 0.79±0.04     & 0.71±0.05      & 0.69±0.07      & 0.59±0.08     \\ \hline
		MMNet18                                & 0.92±0.02      & 0.79±0.07      & \textbf{0.88±0.06}      & \textbf{0.82±0.08}     \\
		MMNet34                                & \textbf{0.94±0.02}     & \textbf{0.85±0.08}      & 0.86±0.10      & 0.81±0.13     \\ \hline
	\end{tabular}
\end{table}

\subsubsection{PGT Dataset}
Table \ref{tab1} shows the comparison between MMNet and other models in the parotid tumor dataset. The experimental results show that our model has a total accuracy of 90.0\% and a standard deviation of 0.032, which is far ahead of the most advanced methods, and has fewer parameters and fast operation speed. Compared with 3D-Resnet34, MMNet34 has fewer parameters (52M vs 64M), higher performance (90.0\% vs 73.3\%), and nearly four times the operation speed (0.72 vs 2.7). Moreover, MMNet has higher precision and lower standard deviation in almost every classification, which reflects that it has stronger robustness in smaller medical image datasets. In comparison with HyperdenseNet, our method makes use of the flow of attention weight to improve the performance of the model. This also shows that for medical images, a large number of model parameters may not bring better results. We also compare it with MedicalNet and Model Genesis, which shows that large-scale medical data pre-training may not bring greater improvement than ImageNet.

\subsubsection{MRNet Dataset}
The experimental results of different models are summarized in Table \ref{tab2}. The ROC curves for MMNet18 and MMNet34 is shown in Fig. \ref{fig_7}. On MRNet, MMNet leads the other state-of-the-art implementations in almost all metrics. Specifically, MMNet18 leads in 3 out of 12 metrics, MMNet34 leads in 8 metrics, and only ELNet slightly outperforms MMNet18 and MMNet34 in the sensitivity of abnormality diagnosis. it is worth mentioning that our method outperforms existing methods by 10.8\% in the sensitivity of ACL diagnosis, which well demonstrates the good robustness of MMNet in the diagnosis of minor injuries. In addition, compared to ELNet and MRPyrNet, our method does not rely on any domain knowledge of knee injuries. For example, ELNet requires radiologists to select the slice in the MRI sequence that is most likely to contain pathological information, whereas MRPyrNet always assumes that the abnormality is always present in the center of the MRI sequence. While these 2D slice-based models accomplish the detection of abnormalities with the help of the prior knowledge, our implementation can extract common abnormal patterns in multiple MRI sequences more efficiently and robustly through the hybrid dimensional deep model and a dynamic mutual attention module to make a more accurate diagnosis.

\subsubsection{PROSTATEx Dataset}
\begin{figure*}[!t]
	\centering
	\includegraphics[width=7in]{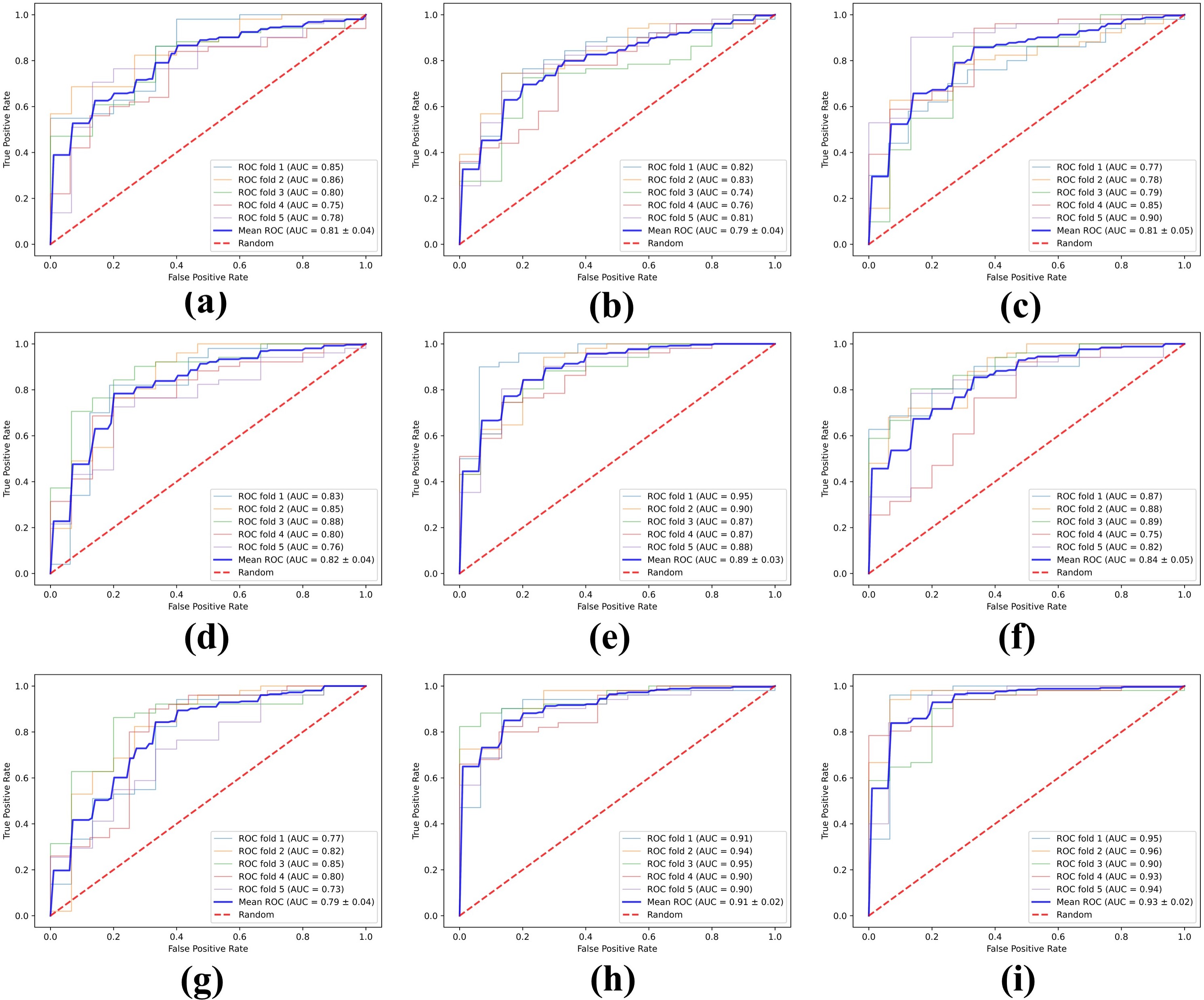}
	\caption{The ROC curves for each experiment in the PROSTATEx dataset. (a) to (i) are arranged in the same order as in Table \ref{tab3}}
	\label{fig_8}
\end{figure*}

We further compared our method with the state-of-the-art methods on the PROSTATEx dataset, and the results are shown in Table \ref{tab3}. The ROC curves of different methods are shown in Fig. \ref{fig_8}. Our smaller model achieved a ROC-AUC of 0.92, a sensitivity of 0.79, a specificity of 0.88, and an accuracy of 0.82. The larger model achieves a ROC-AUC of 0.94, a sensitivity of 0.85, a specificity of 0.86, and an accuracy of 0.81, which are much better than the most recent existing methods. In particular, our model outperforms the MISN network designed for the PROSTATEx dataset. Compared with this state-of-the-art approach, we use a hybrid dimensional model and attention mechanism to extract features effectively and obtain higher performance. In addition, our method is also superior to other 3D CNN-based methods. Moreover, it greatly surpasses other methods that use pre-trained models.

\subsection{Ablation Study}
In this section, we performed ablation analysis on SAM and MAF, to validate the effectiveness of our proposed model. Next, we discussed the impact of the value of $\alpha$ in the MAF.

\begin{table}[]
	\centering
	\caption{Performance comparison between different strategies.}
	\label{tab4}
	\renewcommand\arraystretch{1.3}
	\begin{tabular}{lllll}
		\hline
		Methods & HDB & SAM & MAF & Mean Acc \\ \hline
		MMNet34 & \checkmark   &     &     & 83.6±4.6 \\
		MMNet34 & \checkmark   & \checkmark   &     & 86.4±3.0 \\
		MMNet34 & \checkmark   & \checkmark   & \checkmark   & \textbf{90.0±3.2} \\ \hline
	\end{tabular}
\end{table}

\subsubsection{Ablation Experiments on SAM and MAF}
Table 2 shows the ablation analysis results of key modules in MMNet. The MMNet using only hybrid dimensional blocks has an accuracy of 83.6\% and a standard deviation of 0.046. The model with SAM has an accuracy of 86.4\% and a standard deviation of 0.03. The model with both SAM and MAF has the best performance, with an accuracy of 90.0\% and a standard deviation of 0.032.

\begin{table}[]
	\centering
	\caption{Performance comparison between the different values of $\alpha$.}
	\label{tab5}
	\renewcommand\arraystretch{1.3}
	\begin{tabular}{lllllll}
		\hline
		The value of $\alpha$    & 0     & 0.2   & 0.4   & 0.6   & 0.8   & 1     \\ \hline
		Mean Acc & 82.34 & 87.84 & 87.23 & \textbf{90.02} & 88.86 & 86.39 \\ \hline
	\end{tabular}
\end{table}

Experiments show that SAM and MAF have different contributions to the model. In addition, MMNet, which only has HDB, is also ahead of P3D in terms of indicators, showing the advantages of our hybrid dimensional structure in capturing features in 3D medical images.

\subsubsection{Ablation Experiments on $\alpha$ value of MAF}
In order to get the influence of self-attention weight in MAF on the model, we adjust the $\alpha$ value of self-attention weight in Eq. 11 while keeping other experimental conditions unchanged, and get the results as shown in Table \ref{tab5}. When $\alpha = 0$, that is, the attention weight is completely exchanged between modalities, and the average accuracy is the lowest, which is 82.34\%. With the increase of $\alpha$, the accuracy showed a trend of first rising and then declining. When the self-attention weight is 0.6, the average accuracy is the highest, which is 90.02\%. At this time, the model can focus on complementary information in cross-modal. When the self-attention weight is 1, the average accuracy is 86.39\%, which is better than $\alpha = 0$. This shows that excessive exchange of cross-modal information will reduce the ability of the model to find complementary information.

\begin{figure}[!t]
	\centering
	\includegraphics[width=3.3in]{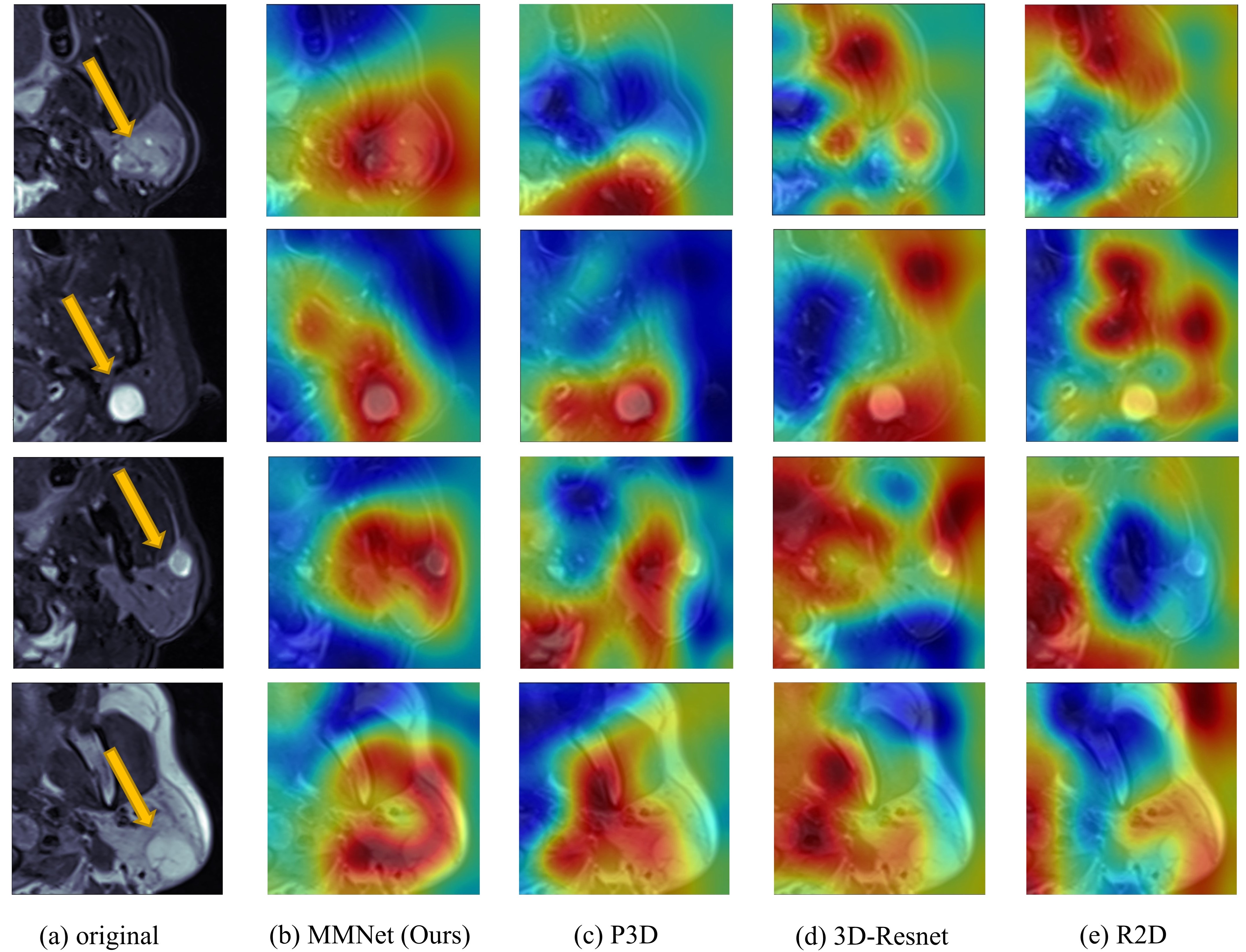}
	\caption{Visualization of Class Activation Mapping for several models. (a) represents some 2D slices of the original image, and (b) to (e) are visualization of MMNet, P3D, 3D-Resnet and R2D, respectively.}
	\label{fig_6}
\end{figure}

\subsection{Visualization of Class Activation Mapping}
Class Activation Mapping (CAM) \cite{CAM} is a tool for analyzing classified hidden information. It uses the global average pooling before the full connection layer of the model to get the weight map of the image. Some part of the image is the main basis for the model to make classification decisions, and CAM can well visualize these areas. As shown in Fig. \ref{fig_6}, we show some 2D slices of the image.

We can find that MMNet has a higher response in capturing pathological regions, which indicates that our proposed model can deal with objects with different shapes and structures. Also, R2D aliases channels and depth at the same time, resulting in a significant loss of 3D information of the image, which leads to its poor performance in capturing pathological regions. This shows the importance of maintaining the consistency of channel and depth in 3D medical images.

\section{Conclusion}
This paper discusses a novel mutual attention-based hybrid dimensional network, MMNet, which utilizes multimodal 3D medical images to boost the performance of lesion detection models. Experiments show that our method can substantially improve the performance of medical image recognition. This method adopts a novel hybrid dimension architecture, and integrates the features of two CNN to simulate the cross-modal dependency, so as to achieve high performance and high efficiency. In summary, our method improves the ability of feature extraction and provides great advantages for multimodal 3D medical image recognition.

\bibliographystyle{IEEEtran}
\bibliography{IEEEexample}
 
\end{document}